\documentclass[sigconf]{acmart}

\AtBeginDocument{%
	\providecommand\BibTeX{{%
			\normalfont B\kern-0.5em{\scshape i\kern-0.25em b}\kern-0.8em\TeX}}}

\acmYear{2022}\copyrightyear{2022}
\setcopyright{rightsretained}
\acmConference[PACT '22]{International Conference on Parallel Architectures and Compilation Techniques}{October 10--12, 2022}{Chicago, IL, USA}
\acmBooktitle{International Conference on Parallel Architectures and Compilation Techniques (PACT '22), October 10--12, 2022, Chicago, IL, USA}
\acmPrice{}
\acmDOI{10.1145/3559009.3569651}
\acmISBN{978-1-4503-9868-8/22/10}

\usepackage{microtype}
\usepackage{graphicx}
\usepackage{subcaption}
\usepackage{outlines}
\usepackage{booktabs} 
\usepackage{amsfonts}

\usepackage{tabularx}
\usepackage[frozencache=true,cachedir=.]{minted}
\usepackage{hyperref}
\usepackage{amsmath}
\usepackage{cleveref}
\usepackage{algorithm}
\usepackage{algorithmic}

\DeclareMathOperator*{\argmin}{arg\,min}

\newcommand{\backend}{{\em backend}}

\newcommand{\ours}{{\em Collage}}

\begin{document}
	
	\title{Collage: Seamless Integration of Deep Learning Backends with Automatic Placement}
	
	\author{Byungsoo Jeon}
	\authornote{Both authors contributed equally to this research.}
	\email{byungsoj@cs.cmu.edu}
	\affiliation{%
		\institution{Carnegie Mellon University}
	}
	
	\author{Sunghyun Park}
	\authornotemark[1]
	\authornote{The work was done during their degree programs. }
	\email{spark@octoml.ai}
	\affiliation{%
		\institution{OctoML}
	}
	
	\author{Peiyuan Liao}
	\email{peiyuanl@andrew.cmu.edu}
	\affiliation{%
		\institution{Carnegie Mellon University\\Praxis Pioneering}
	}

	\author{Sheng Xu}
	\email{jackyxu1997@gmail.com}
	\authornotemark[2]
	\affiliation{%
		\institution{Amazon Web Services}
	}
	
	\author{Tianqi Chen}
	\email{tqchen@cmu.edu}
	\affiliation{%
		\institution{Carnegie Mellon University\\OctoML}
	}
	
	\author{Zhihao Jia}
	\email{zhihao@cmu.edu}
	\affiliation{%
		\institution{Carnegie Mellon University}
	}

	\renewcommand{\shortauthors}{Byungsoo and Sunghyun, et al.}
	
	\begin{abstract}
	The strong demand for efficient and performant deployment of Deep Learning (DL) applications prompts the rapid development of a rich DL ecosystem.
	To keep up with this fast advancement, it is crucial for modern DL frameworks to efficiently integrate a variety of optimized tensor algebra libraries and runtimes as their \textbf{backends} and generate the fastest possible executable using these backends.
	However, current DL frameworks require significant manual effort and expertise to integrate every new backend while failing to unleash its full potential. Given the fast-evolving nature of the DL ecosystem, this manual approach often slows down continuous innovations across different layers; it prevents hardware vendors from the fast deployment of their cutting-edge libraries, DL framework developers must repeatedly adjust their hand-coded rules to accommodate new versions of libraries, and machine learning practitioners need to wait for the integration of new technologies and often encounter unsatisfactory performance.
	
	
	
	In this paper, we propose \ours{}, a DL framework that offers seamless integration of DL backends. \ours{} provides an expressive backend registration interface that allows users to precisely specify the capability of various backends.
	By leveraging the specifications of available backends, \ours{} automatically searches for an optimized backend placement strategy for a given workload and execution environment. 
	Our evaluation shows that \ours{} outperforms the best existing framework for each hardware by $1.26\times$, $1.43\times$, $1.40\times$ on average on NVIDIA's RTX 2070 GPU, V100 GPU, and Intel's Xeon 8259CL CPU, respectively. 
	Collage has been open-sourced~\footnote{\href{https://github.com/cmu-catalyst/collage}{https://github.com/cmu-catalyst/collage}} and deployed in Apache TVM.
	
	
	
\end{abstract}

	
	
	
	
	\maketitle
	


\section{Introduction}

Due to the explosive popularity of Deep Learning (DL) applications, there are tremendous demands for performant and efficient software/hardware stacks for DL computations. These strong demands have driven both industry and academia to invest a significant amount of effort in developing various hardware devices~\cite{TPU, ANE, NVDLA}, software libraries~\cite{cuDNN, OneDNN, TensorRT, OpenVINO, khan2019miopen}, compilers~\cite{Halide_AutoSched, baghdadi2019tiramisu, Ansor, kjolstad2017tensor, phothilimthana2021flexible, ma2020autohoot, chelini2020automatic, rasch2019generating, bastoul2022optimizing, chandrasekhar2019igc, lueh2021c}, and DL frameworks~\citep{PyTorch, TensorFlow, XLA, rotem2018glow, cyphers2018intel, ragan2013halide}. Both the hardware and software stacks for DL have been diversified, resulting in a rich and fast-evolving ecosystem.


Within this ecosystem, today's DL frameworks can leverage a variety of optimized software libraries~\cite{cuDNN,wang2014intel} and runtimes~\cite{TensorRT, khan2019miopen} as their \textbf{backends}~\footnote{We define a {\em backend} as a kernel library or a runtime framework that takes DL workloads as inputs and provides an optimized low-level target code.} to deliver fast execution. Existing backends can be grouped into two categories based on their capabilities. First, {\em operator kernel libraries}~\citep{cuDNN, wang2014intel, khan2019miopen} provide efficient low-level kernel API for individual DL operators (e.g., convolution). These libraries often support {\em operator fusion}, which combines multiple operators into a single kernel based on certain fusion rules~(e.g., cuDNN fusion engine)~\cite{chen2018tvm, cuDNN, niu2021dnnfusion, ma2020rammer, zheng2020fusionstitching, elgamal2017spoof}.
Second, {\em graph inference libraries}~\cite{TensorRT, OneDNN} take an entire DL model as input and produce efficient run-time code. In addition to the optimizations that operator kernel libraries provide, the graph inference libraries also consider graph-level cross-kernel optimizations, such as memory optimizations~\cite{trt_opt}.

\begin{figure}[t]   
    \centering
   \includegraphics[width=1.0\linewidth]{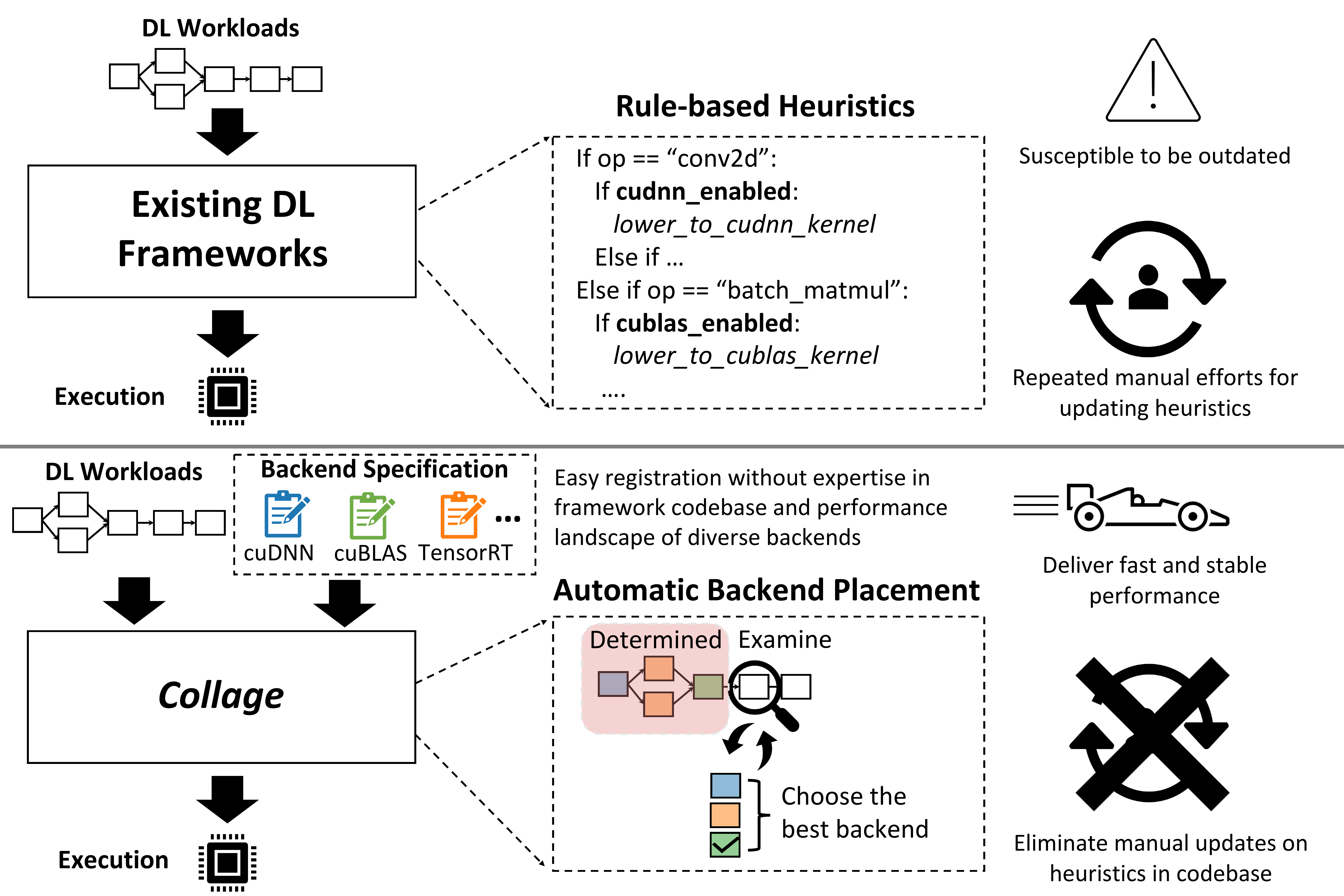}
    \caption{A comparison between existing DL frameworks and \ours{}. Existing frameworks (top) use rule-based heuristics to integrate different backends. In contrast, \ours{} provides an automatic search algorithm to find optimized placement of backends for a given hardware platform. New backends can be easily integrated into \ours{} through the backend registration interface.} 
    \label{fig:comparison}
\end{figure}

There are strong demands for high-performance DL backends in both industry and academia. However, seamless integration of diverse and rapidly advancing DL backends requires addressing two key challenges: (1) incorporating a wide variety of available backends with different programming models and performance characteristics, and (2) optimizing placement of backends to effectively assign DL computations to various backends by leveraging the performance advantages of each backend. We refer to this overall problem as \textbf{backend integration problem}.

To deal with the backend integration problem, existing DL frameworks~\cite{TensorFlow,PyTorch} rely on rule-based heuristics manually designed by experts~(Figure~\ref{fig:comparison}). 
These heuristics often directly offload the entire workload to a single backend (e.g., TensorRT) whenever applicable.
Otherwise, DL frameworks lower individual operators to different backends based on a fixed priority-based strategy; for example, in PyTorch, cuDNN has the highest priority for convolution, while cuBLAS is the first choice for matrix multiplication.

\begin{figure}[!t]
	\centering
	\includegraphics[width=1.0\linewidth]{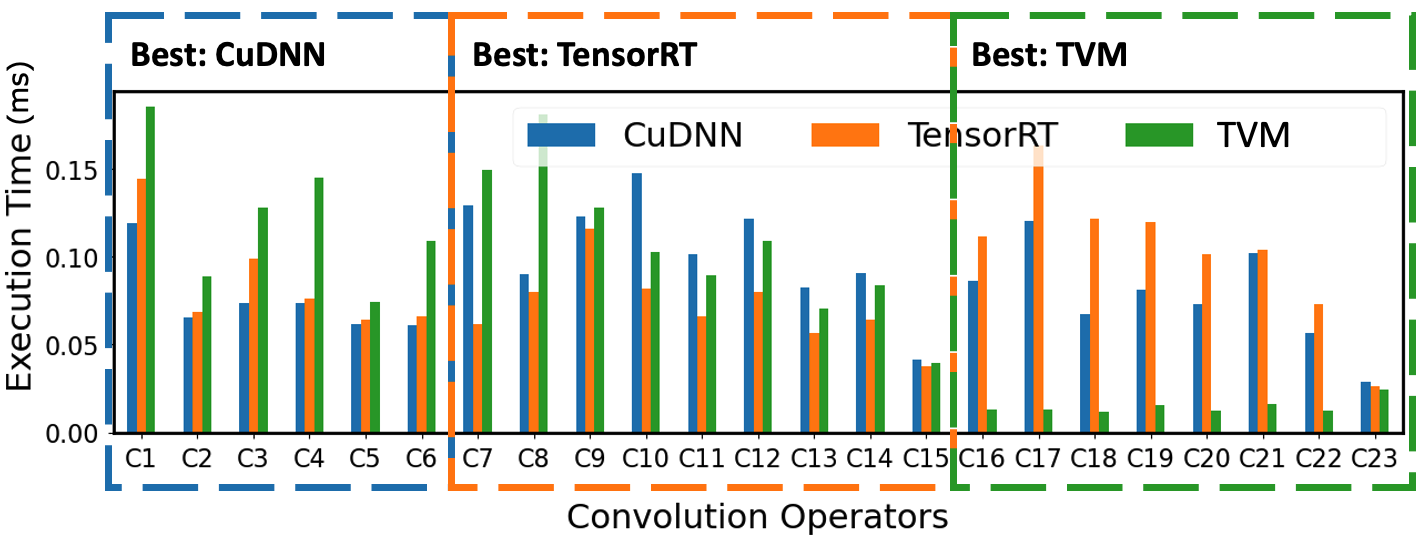}
	\caption{Performance of various convolutions (C\#) with different configurations (e.g., input tensor shape, kernel size) in ResNext-50 on NVIDIA RTX 2070; Note that there is no single \backend{} that is the best for all convolutions. }
	\label{fig:perf_diveristy}
\end{figure}

However, even for the same type of operators, the optimal backend varies depending on the hardware (e.g., different types of GPUs) and operator configuration~(e.g., tensor shape, padding) as depicted in \Cref{fig:perf_diveristy}. 
As a result, the hand-coded heuristics in current DL frameworks may leave substantial performance on the table. Besides, existing frameworks require significant expertise in both framework and performance landscape of diverse backends as developers need to directly modify the complex lowering heuristics (e.g., more than ten thousand lines of code in PyTorch) in a framework to introduce a new backend or reflect any backend updates. These handcrafted heuristics are hard to maintain and keep up with the rapid developments in backends. This is a major bottleneck for various machine learning personas, since the integration workflow requires repetitive manual efforts to accommodate new backends. This integration overhead hinders hardware vendors from deploying their cutting-edge libraries and delays machine learning practitioners from employing newest system-level supports.

In this paper, we aim to design a system that can provide seamless backend integration workflow with high performance. Building such a solution requires addressing two key challenges.
First, it is non-trivial to integrate diverse backends with different characteristics into a system while maintaining their full capabilities. Often times, backend capability is intricate to capture accurately since today's DL backends generally support sophisticated operator fusion with various constraints (e.g., fusing convolution ops with 3x3 kernel). Second, the search space of backend placement is extremely large, whose size grows exponentially in the number of operators in a DNN and the number of available backends.
The search space is also highly irregular due to diverse backend capabilities and operator fusion patterns.

In \ours{}, we advocate for a new approach to tackling these challenges, as shown in the bottom of \Cref{fig:comparison}.
\ours{} contains two key components. First, to integrate diversified backends, \ours{} provides a descriptive {\em backend registration interface} to specify a backend's capability based on its supported operator type (e.g., conv), configurations (e.g., kernel size), and its fusion rule. This interface only requires basic understanding of our pattern language and backend capability in contrast to existing frameworks that require considerable expertise in both the performance landscape of varied backends and the coding skills for backend placement rules in existing frameworks. \ours{} allows easy backend registration for a new backend (e.g., ~100 LoC for all possible operators) or a new operator pattern support (e.g., 1 LoC in most cases).
Second, to efficiently optimize backend placement, \ours{} employs a {\em two-level optimization} to deal with unique chacteristics of two backend categories (i.e., operator kernel library and graph inference library). Our system automatically explores possible matches between an input computation graph and backend operator patterns to find optimized placements by taking available backends and an underlying hardware into consideration.

To sum up, \ours{} significantly lowers the bar in the current backend integration workflow by eliminating the need to modify the placement heuristic. With simple registration from users, \ours{} can immediately launch the automatic placement optimizer without any intricate manual consideration for the capability of new backend and its performance relation with other backends across different workloads and hardware architectures.

\begin{figure*}[!t]
	\centering
	\includegraphics[width=2\columnwidth]{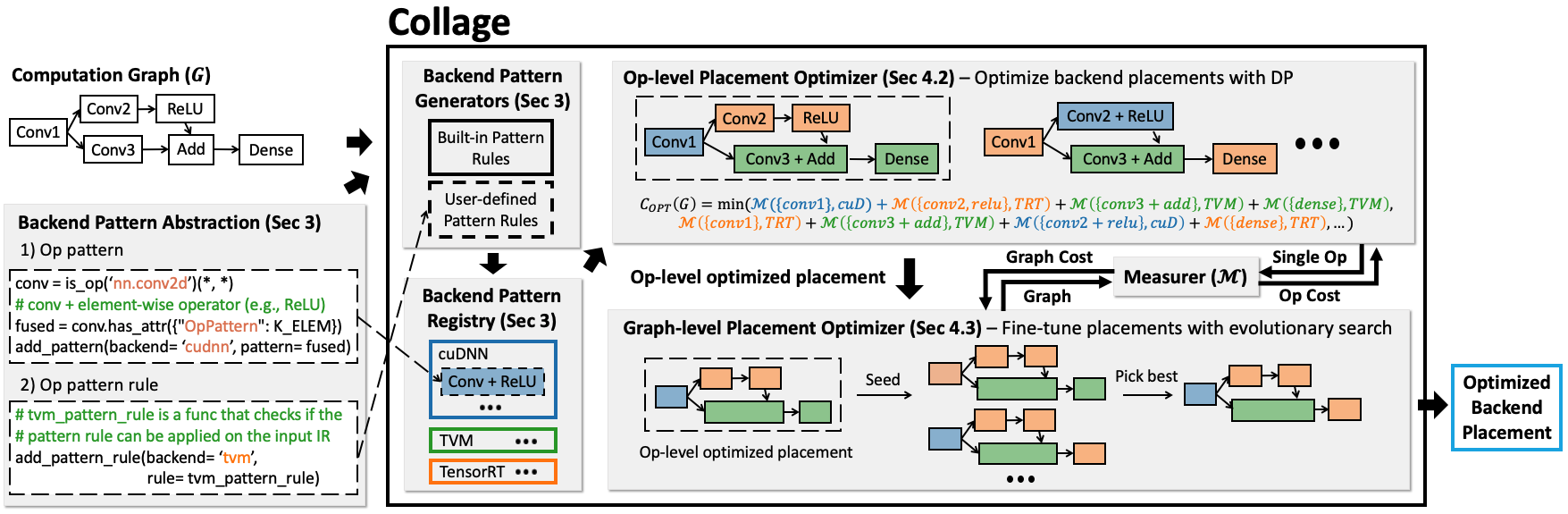}
	\caption{System overview of \ours{}. By using our backend specification interface, users can efficiently register diverse backend patterns supported by diverse backends. Then, with its two-level optimization process, \ours{} automatically optimizes backend placement for an underlying execution environment.}
	\label{fig:instance_level_overview}
\end{figure*}

This paper makes the following contributions:
\begin{itemize}
    \item We identify system and optimization challenges in integration of diversified DL backends and build \ours{} to tackle these challenges.
	
	
	\item We provide a pattern-based interface for quick registration of various backends and their updates with significantly less user efforts and expertise in performance landscape of varied backends and the placement heuristic in the framework codebase.
	
	\item We develop a two-level search method to automatically optimize placement of diverse backends for a given hardware.
\end{itemize}

Our evaluation shows that \ours{} stably outperforms existing DL frameworks across a variety of models and hardware architectures by effectively mix-using multiple backends with their own unique strengths. On average, \ours{} brings $1.26\times$, $1.43\times$, and $1.40\times$ speedup on two different NVIDIA GPUs and an Intel CPU respectively, compared to the best framework for each hardware.


\section{Overview}
\label{sec:overview}
Figure~\ref{fig:instance_level_overview} illustrates the overarching design of \ours{}, which takes a DNN model and the specifications of available backends as inputs, and optimizes backend placement for the underlying hardware. Note that \ours{} considers different sets of backends based on a given target environment (e.g., Intel CPU, NVIDIA GPU) and reflects performance characteristics of backends via the measurer component ($\mathcal{M}$). \ours{} consists of two key components. 

\textbf{Backend pattern abstraction.}
Existing backends provide a variety of programming models for performing DL computations. To decouple backend capability from the placement algorithm and eliminate the manual effort for backend integration, we introduce {\em backend pattern}, a new abstraction for capturing the capability of varied backends. Specifically, a backend pattern defines a set of operators and their possible fusion combinations (e.g., Conv+ReLU) that can be deployed on each backend.
Based on this pattern abstraction, \ours{} provides a straightforward interface to register a backend and specify supported operator patterns. 


Accurate specification is crucial to leverage the full capability of diverse backends. To achieve this goal, \ours{} offers two levels of abstraction. For simple patterns, \ours{} allows users to enumerate the supported operator patterns. However, this approach may not cover the full capability of backends with advanced operator fusion engines~\cite{chen2018tvm, niu2021dnnfusion, cuDNN, TensorRT}. To enable more flexible specification, \ours{} also allows users to bring their pattern rules that specify supported operator kinds and complex operator fusion rules. When those rules are provided, the {\em pattern generator} automatically identifies all legitimate operator fusion patterns on a given computation graph and adds them into the backend pattern registry. \S\ref{sec:pattern_abstract} provides details. 

\textbf{Backend placement optimizer.} 
Once all available patterns are registered in the pattern registry, \ours{} uses a {\em two-level optimization} approach to discovering an optimized backend placement strategy for a given execution environment. As existing operator libraries offer operator-level point of view while graph inference libraries additionally apply cross-kernel optimizations, \ours{} takes two different optimization strategies to exploit their differences.
First, the {\em op-level placement optimizer} explores promising candidates for individual operators, without considering cross-kernel optimizations. By adopting a Dynamic Programming (DP) algorithm, the op-level placement optimizer can efficiently find an optimized backend placement strategy within a minute.
Second, the {\em graph-level placement optimizer} fine-tunes the optimized backend placement using evolutionary search~\cite{fortin2012deap}. This approach compensates for the missing opportunities from the op-level placement optimizer by examining the impact of cross-kernel optimizations.
\S\ref{sec:tuner} discusses the two optimizers in detail.

	\section{Backend Pattern Abstraction}
\label{sec:pattern_abstract}

\usemintedstyle{manni}
\begin{listing}[!t]
\begin{minted}[escapeinside=||,linenos,numbersep=4pt,frame=lines,fontsize=\scriptsize]{python}
import collage

# [Method 1] Explicit pattern specification |\label{line:pat:explicit_begin}|
# Pattern language to describe conv2d + add + relu.
conv = is_op('conv2d')(wildcard(), wildcard()) |\label{line:pat:explicit_0}|
conv_constr = conv.has_attr({"data_layout": "NCHW"}) |\label{line:pat:explicit_1}|
conv_add = is_op('add')(conv_constr, wildcard())  |\label{line:pat:explicit_2}|
conv_add_relu = is_op('relu')(conv_add) |\label{line:pat:explicit_3}|

# Introduce new backend pattern to Collage.
collage.add_backend_pattern(backend='cuDNN', 
                            pattern=conv_add_relu) |\label{line:pat:explicit_end}|

# [Method 2] Pattern rule specification |\label{line:pat:rule_begin}|
class MyPatternRule(collage.BasePatternRule):
   # Define variables
   kFusable = 0
   kElemwise = 1
   # ...
   # Checker for the supported operators. |\label{line:pat:rule_valid_ops_begin}|
   @staticmethod
   def op_rule(op): 
      if op.name == "dense":
         # Dense operator is always supported. 
         return True
      elif op.name == "conv2d":
        # constraints can be verified as well.
        return op.attr["data_layout"] == "NCHW"
      # ... rest of the op rule ...
      return False  |\label{line:pat:rule_valid_ops_end}|
     
   # Checker for fusion patterns.  |\label{line:pat:rule_fusion_begin}|
   #   -- cur_type: type of current fusion group
   #   -- src: seed operator node
   #   -- sink: post-dominator of src
   @staticmethod
   def fusion_rule(cur_type, src, sink): 
      # If current fusion group contains 
      #   at least one conv/matmul (kFusable)
      if cur_type == MyPatternRule.kFusable:  |\label{line:pat:rule_2}|
        # Helper functions can be defined.
        def fchecker(node_pattern): |\label{line:pat:rule_3}|
           return (node_pattern == MyPatternRule.kElemwise) 
        # Check if every operator between src and sink.
        # Helper function can be passed as a checker.
        if collage.check_path(src, sink, fchecker)): |\label{line:pat:rule_4}|
           return True |\label{line:pat:rule_5}|
      # ... rest of the fusion rule ... 
      return False |\label{line:pat:rule_fusion_end}|
     
# Introduce new pattern generation rule to Collage.
collage.add_backend_pattern_rule(backend='TVM',
                                 pattern_rule=MyPatternRule()) |\label{line:pat:rule_end}|
\end{minted}
\captionof{listing}{Example of the backend registration interface. To register a new backend, users can directly enumerate patterns or write a pattern rule that consists of valid operator checker and fusion rule in Python classes.}
\label{code:abstraction}
\end{listing}



As an important component of DL ecosystem, there are diverse fast-evolving DL backends with different programming models and performance characteristics. Depending on their target hardware and design principles, each backend has its own unique strength and coverage. In addition, many backends support various complex operator fusion rules~\cite{cuDNN, chen2018tvm, TensorRT, chen2018tvm, niu2021dnnfusion}, which add significant complexity in their integration with the full capability. Under the hood, existing operator fusion engines often fuse operators based on heuristic fusion rules that examine the type of each operator and the relationship between different types. For instance, a fusion engine may combine multiple operators across different branches into a single kernel as long as they satisfy its fusion rule.

For an adoption of various backends, our system provides two levels of abstraction: {\em pattern} and {\em pattern rule}. Pattern is a direct way to specify all supported operator patterns in \ours{}'s pattern language,  which extends the Relay pattern language~\cite{roesch2018relay}. However, supported patterns can be too complicated to explicitly specify. To incorporate sophisticated patterns, pattern rules offer an expressive way to specify a valid set of operator fusion rules in the form of Python; users can use any Python features to describe complex fusion algorithms. Each pattern rule is used to generate valid patterns for the input workload with our automatic pattern generator. With two levels of abstraction, users can easily incorporate an additional backend by specifying its patterns and pattern rules with an intuitive programming interface. By default, \ours{} provides built-in patterns and pattern rules for popular backends \citep{cuDNN, cuBLAS, TensorRT, wang2014intel, chen2018tvm}.
 
 Listing~\ref{code:abstraction} presents an example of use-case scenarios. If a backend only supports a few simple patterns, users may enumerate those patterns and add them directly to the backend pattern registry~(line~\ref{line:pat:explicit_begin}-\ref{line:pat:explicit_end}). Users can easily check the operators ~(line~\ref{line:pat:explicit_0}), their configurations such as data layouts and kernel sizes~(line~\ref{line:pat:explicit_1}), and the the relationship between operators~(line~\ref{line:pat:explicit_2}-\ref{line:pat:explicit_3}). A wildcard operator is a special placeholder that matches any operator. 
 
 To fully support advanced backends~\cite{chen2018tvm, niu2021dnnfusion, cuDNN, TensorRT}, users can bring their pattern rules to incorporate more complicated patterns with \ours{}'s pattern generator~(line~\ref{line:pat:rule_begin}-\ref{line:pat:rule_end}). To use this feature, users need to provide operator checkers with their potential constraints~(line~\ref{line:pat:rule_valid_ops_begin}-\ref{line:pat:rule_valid_ops_end}) and a fusion rule~(line~\ref{line:pat:rule_fusion_begin}-\ref{line:pat:rule_fusion_end}) in the form of Python methods. Then, the automatic pattern generator in \ours{} will search for valid operator patterns satisfying these rules and add them to the backend pattern registry before optimizing backend placement. 
 
 \begin{figure}[t]
	\centering
	\includegraphics[width=1.0\columnwidth]{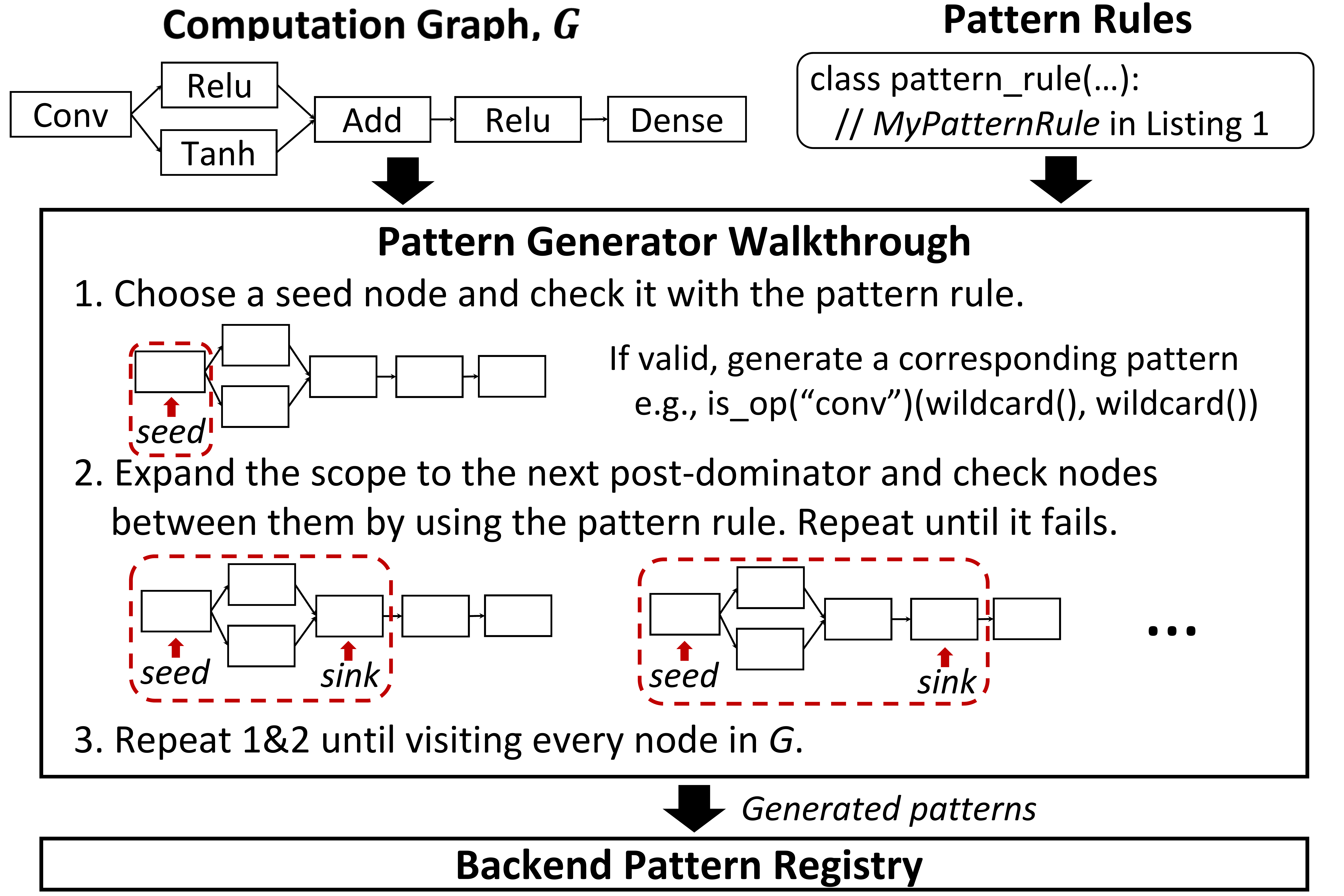}
	\caption{Example illustrating how the backend pattern generator would automatically generate valid patterns with the pattern rule presented in Listing~\ref{code:abstraction}.}
	\label{fig:pattern_generator}
\end{figure}
 
 Figure~\ref{fig:pattern_generator} exhibits how our pattern generator searches for legitimate patterns using given pattern rules on an input computation graph. By visiting every operator in an input computation graph, the pattern generator investigates how far a pattern can grow without breaking the pattern rule. For each operator, the pattern generator first validates whether the operator can be executed on a backend~(line~\ref{line:pat:rule_valid_ops_begin}-\ref{line:pat:rule_valid_ops_end}). If valid, it enlarges the scope one step further and validates whether a set of operators satisfies the fusion rule~(line~\ref{line:pat:rule_fusion_begin}-\ref{line:pat:rule_fusion_end}). For instance, line~\ref{line:pat:rule_2}-\ref{line:pat:rule_5} specify that the assumed backend can fuse element-wise operators following an operator of type {\tt kFusable}, which includes convolution and matrix multiplication. Whenever a group of operators satisfying the rule is found, the pattern generator produces a corresponding pattern and adds it to the backend pattern registry. Then, it enlarges the scope of interests one step further again to see if a bigger pattern can be found. This approach allows \ours\ to incorporate advanced backends, such as TVM, cuDNN, DNNL and TensorRT, without missing any pattern.




	
\section{Backend Placement Optimization}
\label{sec:tuner}
\subsection{Problem Definition}
  
  \ours{} attacks the backend placement problem to find the best use of available backends and maximize performance. Consider a computation graph $\mathcal{G}$ and a set of backend patterns $\mathcal{B}$ in \ours{}'s backend pattern registry. $\mathcal{G}$ is a Directed Acyclic Graph (DAG) where each node represents a tensor operator (e.g., convolution, matrix multiplication). $b = (p, d)\in\mathcal{B}$ is a pair of an operator pattern $p$ and a backend identifier $d$, such as cuDNN, cuBLAS, etc.
  
  With $M$ matched subgraphs $g_i$ and backend patterns $b_i$ for $i \in \{1, 2, \cdots M\}$, let $\mathcal{P}(\mathcal{G}) = \{(g_i, b_i) | b_i \in \mathcal{B}, \bigcup_{i=1}^{M} g_i=\mathcal{G},g_i \cap g_j = \emptyset \text{ for all }i,j \in \{1,2, \cdots, M\} \text{ where } i \neq j \}$ be a backend placement strategy on a computation graph $\mathcal{G}$ and $Cost(\mathcal{P}(\mathcal{G}))$ be the execution time of a placement $\mathcal{P}(\mathcal{G})$. In this work, we aim to find a backend placement strategy $\mathcal{P}_{opt}$ that minimizes $Cost(\mathcal{P}(\mathcal{G}))$.
  This problem can be formalized as follows:
  \begin{equation} \label{eq:goal}
     \mathcal{P}_{opt}(\mathcal{G}) = \argmin_{\mathcal{P}(\mathcal{G})} Cost(\mathcal{P}(\mathcal{G}))
  \end{equation}



\subsection{Op-level Placement Optimizer}
\label{sec:op-lv-tuner}
To efficiently evaluate numerous candidates with different placement and prune the search space, \ours{} conducts an op-level placement optimization as the first step. Its goal is to map all operators on the computation graph to the most efficient set of low-level kernel implementations from available backends fast without considering cross-kernel optimizations in graph inference libraries. As discussed earlier, the graph-level placement optimizer~(\S\ref{sec:graph-lv-tuner}) would make up for the possible performance loss from this simplification. 


With this simplification, low-level kernel executions become independent to each other in a single device execution. Let $s_1$ and $s_2$ be subgraphs of $\mathcal{G}$ where $s_1 \cup s_2 = \mathcal{G}$, $s_1 \cap s_2 = \emptyset$. Then, the following additive relationship~\cite{jia2019taso} between the run-time cost of $\mathcal{P}(s_1)$ and $\mathcal{P}(s_2)$ can be used to determine $Cost(\mathcal{P}(\mathcal{G}))$:
\begin{equation}
\begin{split}
  \label{eq:additive} 
 Cost(\mathcal{P}(\mathcal{G})) = Cost(\mathcal{P}(s_1)) + Cost(\mathcal{P}(s_2)) + \epsilon\\
\end{split}
\end{equation} 
where $\epsilon$ is a context switching cost (e.g, driver overhead), which is nearly constant empirically. Note that \ours{} avoids data transfers between different backends on the same device by only exchanging data pointers to the tensors (e.g., s1 and s2) using the zero-copy mechanism. With this cost model, it is possible to cheaply approximate the cost of a graph by partitioning a graph into smaller subgraphs and summing up their cost.
Despite the efficient cost model, excessively large number of possible placement strategies and a variety of fusion patterns make search non-trivial.

\begin{figure}[t]
	\centering
	\includegraphics[width=1.0\linewidth]{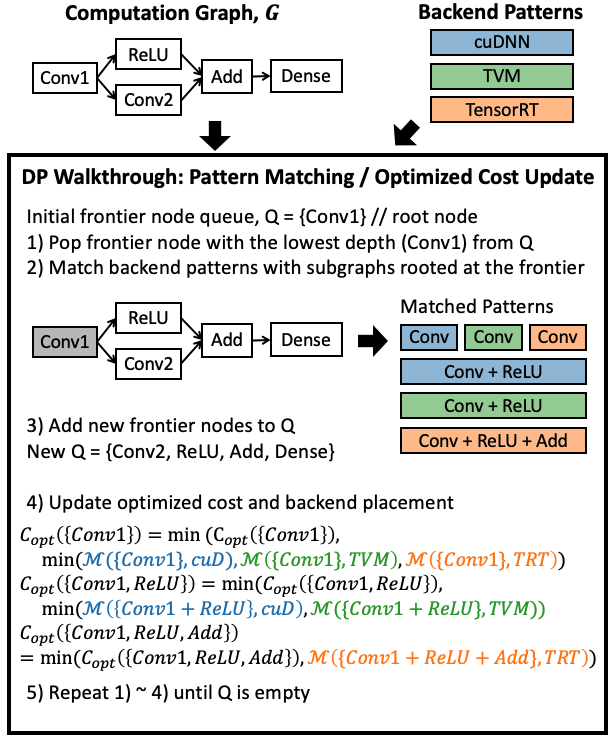}
	\caption{Example of Dynamic Programming (DP) procedures.  By visiting over each frontier node, DP algorithm matches backend patterns and update the optimized placement and its cost. For simplicity, optimized placement update is omitted.}
	\label{fig:dp_example}
\end{figure}
To address this challenge, we propose a Dynamic Programming (DP) method for optimizing backend placement at the operator level. By using the additive relation (Equation~\ref{eq:additive}), we deduce the following recurrence relation of optimized backend placement $\mathcal{P}_{opt}(s)$ and its cost $\mathcal{C}_{opt}(s)$ for any subgraph $s \subset \mathcal{G}$. This breaks down a problem of finding $\mathcal{P}_{opt}(\mathcal{G})$ into smaller problems of finding $\mathcal{P}_{opt}(s)$.
\begin{equation}
    \begin{split}
      \label{eq:dp}
      \mathcal{P}_{opt}(s) &= \mathcal{P}_{opt}(s_{min}) \cup \mathcal{P}(g_{min}) \\
      \mathcal{C}_{opt}(s) &=
      \begin{cases}
        0 &\text{if } s = \emptyset \\
        \mathcal{C}_{opt}(s_{min}) + \mathcal{M}(\mathcal{P}(g_{min})) + \epsilon & \text{otherwise}
      \end{cases} 
    \end{split}
\end{equation}
where $s_{min}$ and $g_{min}$ are
\begin{equation}
    \begin{split}
        \argmin_{s'\cup g'= s, s' \cap g' = \emptyset} \{ \mathcal{C}_{opt}(s') + \mathcal{M}(\mathcal{P}(g')) + \epsilon \}
    \end{split}
\end{equation}
 
 \begin{algorithm} [t]
	\caption{Op-level Placement Optimization: DP} \label{alg:dp}
	\begin{flushleft}
	    \textbf{Input:} Computation graph $\mathcal{G}$ and set of backend patterns $\mathcal{B}$ \\
	    \textbf{Output:} Optimized placement $\mathcal{P}_{opt}(\mathcal{G})$\\ 
    \end{flushleft}
	\begin{algorithmic}[1]
		\STATE // $v_0$: a root of $\mathcal{G}$, $\mathcal{Q}$: a priority queue sorted by node depth
		\STATE $\mathcal{Q} = \{v_0\}$ 
		\REPEAT \label{line:while}
		    \STATE // $v_s$ is a frontier node
    		\STATE $v_s = \mathcal{Q}$.\texttt{dequeue}()
    		\FOR {$ b_i \in \mathcal{B}$} \label{line:for1}
         		\STATE // Find a subgraph $g$ rooted at $v_s$ that matches $b_i$
        		\IF {$g$ = get\_match($v_s$, $b_i$)} \label{line:pattern_match}
        		    \STATE // $\mathcal{F}$ is a set of new frontier nodes after matching
            		\FOR {$ v_j \in \mathcal{F}$}  \label{line:for2}
            		    \IF{$v_j$ has never been added to $\mathcal{Q}$} 
                		    \STATE $\mathcal{Q}$.\texttt{enqueue}($v_j$)
                		\ENDIF
            		\ENDFOR
            		\STATE
                    \STATE // $\mathcal{P}(g)=\{(g,b_i)\}$
                    \STATE // $\mathcal{M}$ is a measurer
            		\STATE // $\mathcal{S}$ is a set of subgraphs, each of which includes all nodes before $v_{s}$ in post-order and does not include $g$
            		\STATE // $\epsilon$ is a constant for context switching cost
            		\FOR {$ s_j \in \mathcal{S}$}  \label{line:for3}
            		    \IF{$\mathcal{C}_{opt}(s_j \cup g) > \mathcal{C}_{opt}(s_j) + \mathcal{M}(\mathcal{P}(g)) + \epsilon$}
                    		\STATE $\mathcal{C}_{opt}(s_j \cup g) = \mathcal{C}_{opt}(s_j) + \mathcal{M}(\mathcal{P}(g)) + \epsilon$
                    		\STATE $\mathcal{P}_{opt}(s_j \cup g) = \mathcal{P}_{opt}(s_j) \cup \mathcal{P}(g)$
                		\ENDIF
            		\ENDFOR
        		\ENDIF
    		\ENDFOR
		\UNTIL $\mathcal{Q} = \emptyset$
		\STATE
		\STATE \textbf{return} $\mathcal{P}_{opt}(\mathcal{G})$
	\end{algorithmic}
\end{algorithm}
 $s'$ represents a subgraph that is already examined while $g'$ is a subgraph that is going to be evaluated with a measurer $\mathcal{M}(\cdot)$, which takes a backend placement strategy and returns its actual run-time cost on the execution environment. We query the measurer at the granularity of a backend pattern that matches with $g'$, which is either single or multiple operators (operator fusion) that will be lowered to a single low-level kernel. This approach ensures that we always measure a single kernel and add it up to compute the cost of larger subgraphs. To avoid the repetitive and expensive measurement overhead~(i.e., compilation + multiple runs on the actual hardware), we cache the result to the log for the future usage. With this approach, we can efficiently explore possible backend placements and evaluate them.



Figure~\ref{fig:dp_example} illustrates an simplified walkthrough example of our DP method. By traversing a computation graph $\mathcal{G}$, it solves smaller problems of finding $\mathcal{P}_{opt}(s)$ for a subgraph $s \subset \mathcal{G}$ and eventually discovers $\mathcal{P}_{opt}(\mathcal{G})$. First, it puts a root node in the priority queue as an initial frontier node; we define a {\em frontier node} as a node that has the lowest depth among unvisited nodes on a path from the root. Then it pops a frontier node with the lowest depth from the queue and examines if any subgraph rooted at the current frontier node can match any valid backend pattern. Once a matching is found, we add new frontier nodes to the priority queue and measure the cost of the subgraph matched with the backend pattern. If a better placement strategy is found, we update the optimized cost and backend placement strategy based on \Cref{eq:dp}. We repeat these steps until the priority queue is empty. Given that graph inference libraries, such as TensorRT, can also provide competitive operator-level implementations~(Figure~\ref{fig:perf_diveristy}), we also include them in the op-level optimization. Algorithm~\ref{alg:dp} formalizes our DP method.

\textbf{Time complexity.} We derive the time complexity of Algorithm~\ref{alg:dp}. Let $N$ be the number of nodes (operators) in computation graph $\mathcal{G}$, $P$ be the average number of backend pattern matches per frontier, $F$ be the maximum possible number of frontiers for a single match, and $S$ be the maximum number of subgraphs in $\mathcal{S}$~(line \ref{line:for3}). In Algorithm~\ref{alg:dp}, the outermost while loop (line \ref{line:while}) takes $\mathcal{O}(N)$ times to traverse each frontier node in $\mathcal{G}$. For each frontier, there can be $\mathcal{O}(P)$ matches~(line \ref{line:for1}-\ref{line:pattern_match}). For each match, the algorithm iterates over its $\mathcal{F}$~(line \ref{line:for2}) and $\mathcal{S}$~(line \ref{line:for3}) and takes $\mathcal{O}(F+S)$. Therefore, the overall time complexity of our op-level placement optimizer is $\mathcal{O}(NP(F+S))$. In all workloads that we have investigated,  $N < 1000, P < 20, F < 10, S < 200$. As a result, our DP method optimizes placement within a minute by effectively pruning candidates.

\subsection{Graph-level Placement Optimizer}

\label{sec:graph-lv-tuner}

As the op-level placement optimization ignores the effect of cross-kernel optimizations (e.g., scheduling and memory optimizations) in  graph inference libraries, \ours{} introduces the graph-level placement optimizer to fine-tune the potentially sub-optimal backend placement strategies from the op-level. To do so, we need to identify additional operators that are not assigned to graph inference libraries but can benefit from cross-kernel optimizations. Once identified, we offload them to graph inference libraries to extract further improvement. However, a key challenge we must address in this approach is deciding which operators to offload to graph inference libraries among a myriad of candidates.. 

To address this challenge, we represent each backend placement strategy by using a sequence of digits. Each digit implies whether to offload to  graph inference libraries. Since our goal is to offload more operators that can benefit from the cross-kernel optimization, we exclude operators already mapped with a graph inference library from this encoding. This straightforward state representation eliminates the complexity from various graph partitions and their topology.

\begin{figure}[t]
	\centering
	\includegraphics[width=1.0\linewidth]{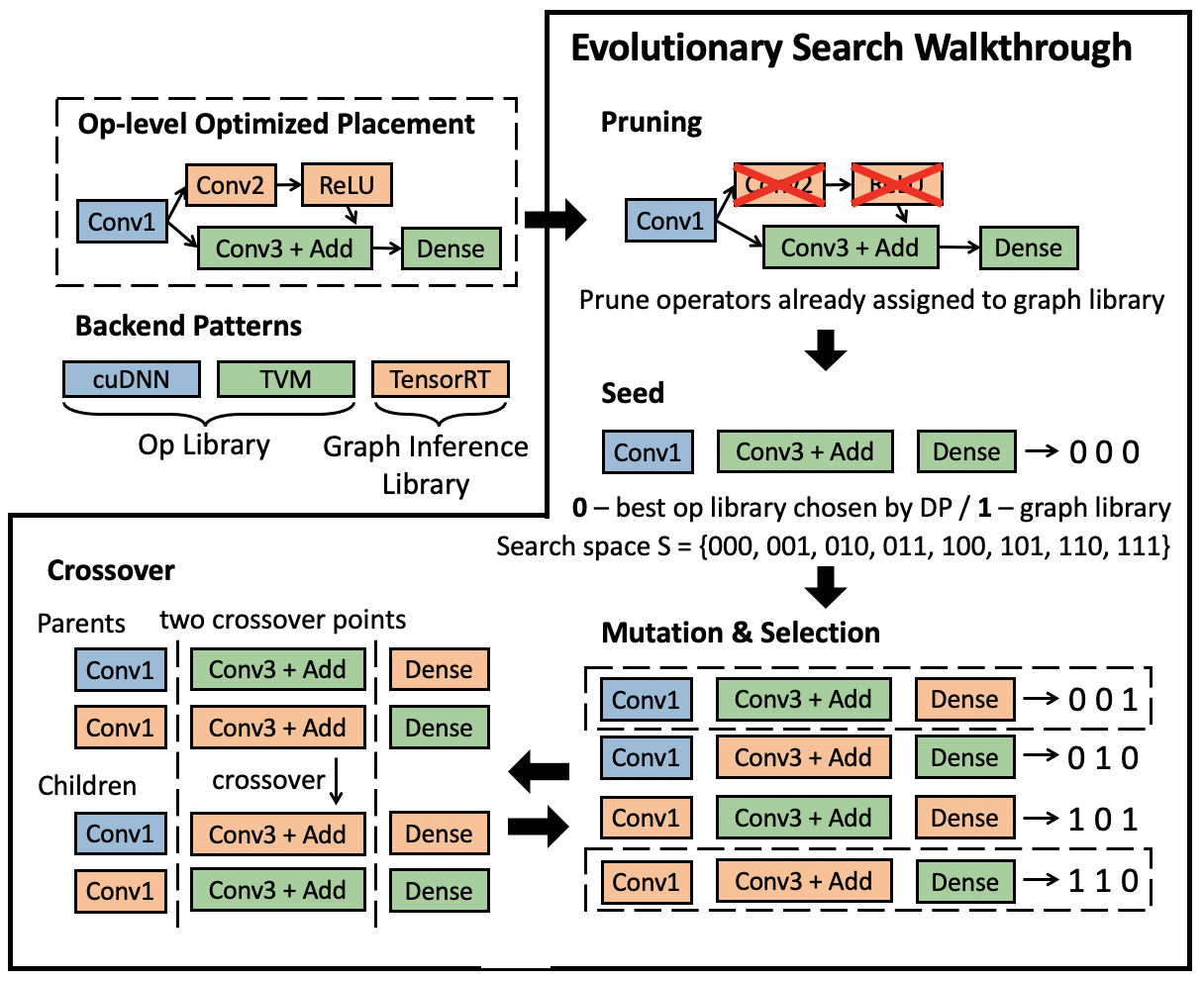}
	\caption{Example of Evolutionary Search (ES) procedure. After pruning search space, it iterates over mutation, selection, and crossover until it reaches saturation or time limit.}
	\label{fig:es_example}
\end{figure}

We adopt an evolutionary search algorithm~\cite{fortin2012deap} for graph-level placement optimization. Figure~\ref{fig:es_example} describes the procedure of our evolutionary search method. For state representation, 0 indicates keeping the decision of the op-level optimizer and 1 means overriding the decision and offloading it to a graph inference library (e.g., TensorRT). To facilitate the search process, we include the op-level optimized placement strategy as one of the seeds to provide a good starting point. The evolutionary algorithm iterates over rounds of mutation, selection, and two-point crossover to fine-tune the backend placement.

\section{Evaluation}
This section aims to answer the following questions:
\begin{itemize}
 \item Can \ours{} effectively optimize real-world DL model execution over diverse backends and target devices compared to the existing DL frameworks? (\S\ref{sec:e2e_eval})
 \item Is optimization time affordable? How much time does each optimization take?  (\S\ref{sec:opt_time})
 \item Does adding more backends improve the performance of \ours{}?  (\S\ref{sec:backend_ablation})
 \item How does backend placement optimized by \ours{} look like?  (\S\ref{sec:backend_case_study})
\end{itemize}

\subsection{Experimental Setup}
\textbf{Implementation.} We built the core of \ours{} in the form of a portable Python library and leveraged diverse backends in different hardware architectures: cuDNN~\cite{cuDNN}, cuBLAS~\cite{cuBLAS}, TVM~\cite{chen2018tvm}, TensorRT~\cite{TensorRT}, MKL~\cite{wang2014intel} and DNNL~\cite{OneDNN}. To orchestrate a runtime execution with multiple backends, Collage uses DLPack to minimize data movement (e.g., tensor) across different backend runtimes by efficiently exchanging pointers of data with zero-copy approach~\cite{DLPack}. Still, even such optimized communications incur certain run-time overhead (e.g., deserialization overhead of the engine in graph inference libraries~\cite{TensorRT-deserialization}). Thus, Collage takes this run-time overhead into account when measuring execution time of various placement candidates. If such run-time overhead is too excessive, Collage will choose another candidate with better performance. 
To leverage full capabilities of backends, their supported patterns and pattern rules are provided based on their official documentation and codebases. Each backend specification with full operator supports only takes about 100 LoC with Collage API.

\textbf{Baselines.} We examine TensorFlow~(TF)~\cite{TensorFlow}, TF-XLA~\cite{XLA}, PyTorch~\cite{PyTorch}, TVM~\cite{chen2018tvm}, and TensorRT~\cite{TensorRT} as DL framework baselines. For TVM, we use AutoTVM to automatically generate the optimized operator schedules for each target. Note that we also integrate TensorRT and TVM as high-performance graph inference libraries in this experiment.

\textbf{Workload.} We evaluate five popular real-world DL inference workloads that cover a wide range of application.
BERT~\cite{devlin2018bert} is a transformer-based language model that achieved the state-of-the-art performance on a spectrum of natural language processing tasks. DCGAN~\cite{radford2015unsupervised} is an extension of the GAN~\cite{goodfellow2020generative} with an unsupervised representation learning mainly for image generation. NasNet-A \cite{nasnet} is one of the most popular machine-generated DL workloads that show strong performance on popular image recognition tasks. 3D-ResNet50 \cite{hara3dcnns} is an extension of widely adopted ResNet50~\cite{he2016deep} for 3D image tasks such as action recognition. ResNeXt50 \cite{Xie_2017_CVPR} introduces a grouped convolution to ResNet50 architecture and improves its model accuracy and computational complexity for image recognition.

Each workload has its own characteristics in terms of its operators and structure. Most of recent models for language application such as BERT are basically a series of the Transformer layers that consist of batch matrix multiplication, layer normalization, softmax, etc. On the other hand, models for vision application such as ResNeXt50 and NasNet-A has a series of layers that has operators including convolutions and non-linear activation functions (e.g., ReLU). In these models, operator configuration (e.g., number of channels and hidden nodes) varies across different layers as you see in \Cref{fig:perf_diveristy}, which leads to performance diversity of DL backends.

\begin{figure*}[t]
	\centering
	\begin{subfigure}[t]{1.0\linewidth}
		\centering
		\includegraphics[width=0.9\linewidth]{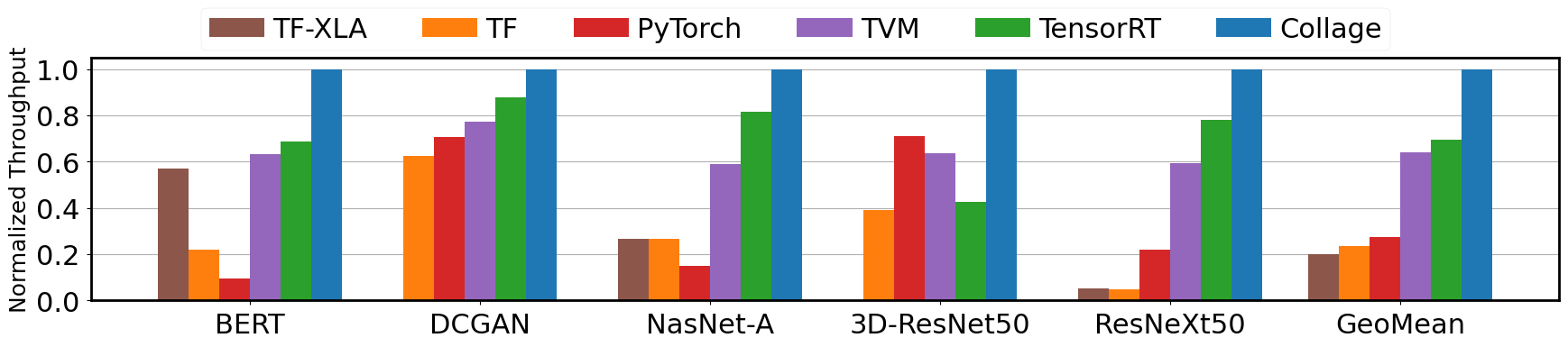}
		\caption{NVIDIA Tesla V100}
		\label{fig:e2e_perf_v100}
	\end{subfigure}
	
	\begin{subfigure}[t]{1.0\linewidth}
		\centering
		\includegraphics[width=0.9\linewidth]{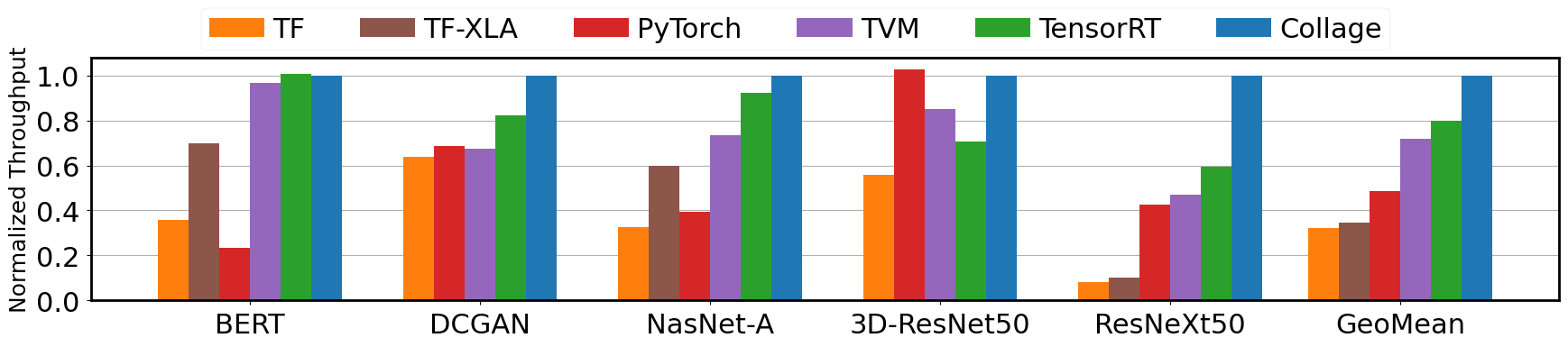}
		\caption{NVIDIA GeForce RTX 2070}
		\label{fig:e2e_perf_rtx2070}
	\end{subfigure}
	
	\begin{subfigure}[t]{1.0\linewidth}
		\centering
		\includegraphics[width=0.9\linewidth]{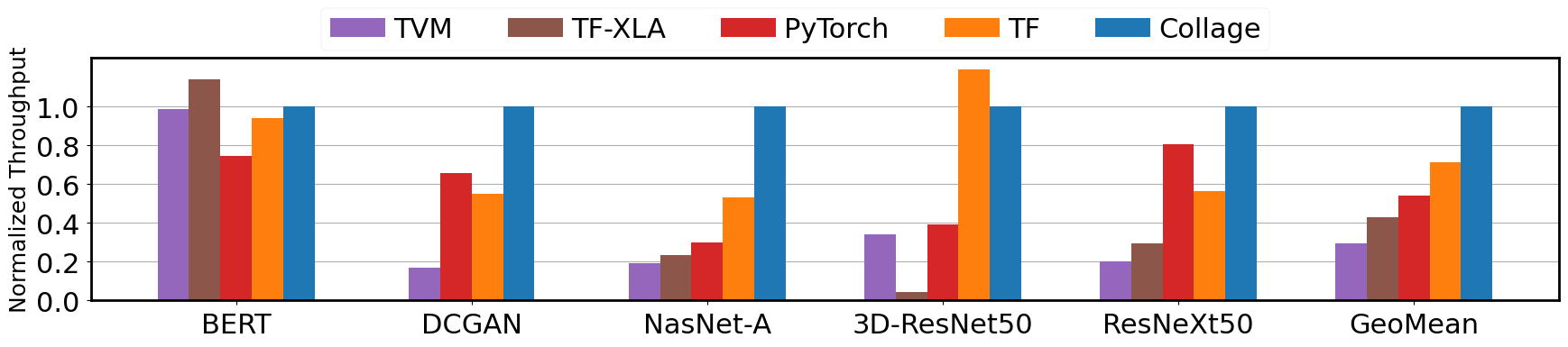}
		\caption{Intel Xeon Platinum 8259CL}
		\label{fig:e2e_perf_xeon}
	\end{subfigure}
	
	\caption{End-to-end performance of state-of-the-arts DL frameworks and \ours{} in five real-life workloads on NVIDIA GPUs and Intel CPU. Throughput of each framework is normalized by the throughput of \ours{}. Following backends are employed for each framework according to target hardware and its capabilities: NVIDIA GPU (cuDNN, cuBLAS, TVM, TensorRT), Intel CPU (MKL, DNNL, TVM).}
	
	\label{fig:e2e_perf_gpu}
\end{figure*}

\subsection{End-to-end Evaluation} \label{sec:e2e_eval}
 


To discuss the effectiveness of our approach, we evaluate the end-to-end performance of \ours{} against the baseline frameworks; note that we omit error bars from our figures because we observe marginal standard deviation (less than 3\%) for all results. Note that the performance of TF-XLA is missing for some pairs of workload and targets (e.g., 3D-ResNet50 and NVIDIA GPU) because it has issues with some 3D convolutions for GPU targets and certain image resizing operators.

\Cref{fig:e2e_perf_v100} and \Cref{fig:e2e_perf_rtx2070} presents the end-to-end normalized throughput of \ours{} and existing DL frameworks on two different NVIDIA GPU architectures, Tesla V100 and GeForce RTX2070. Normalized throughput is the throughput of each framework normalized by the throughput of \ours{}. Overall, \ours{} consistently produces the most efficient executable across different workloads and hardware architectures: In terms of geometric mean, \ours{} outperforms the state-of-the-arts by $1.43\times$ on V100 and $1.26\times$ on RTX 2070, respectively. This improvement comes from \ours{}'s  backend placement optimization that effectively leverages the unique strength of various backends.

Figure~\ref{fig:e2e_perf_xeon} exhibits the experimental results on the Intel CPU. Likewise, \ours{} showcases the most stable performance across different workloads on this Xeon architecture while beating the state-of-the-arts by $1.40\times$ in the geometric mean. However, on BERT and 3D-ResNet50, TF-XLA and TF are faster possibly due to their optimizations customized for Intel CPU such as data layout optimization with non-uniform memory access, which is orthogonal to backend placement.

\begin{figure}[t]
	\centering
	\includegraphics[width=0.99\linewidth]{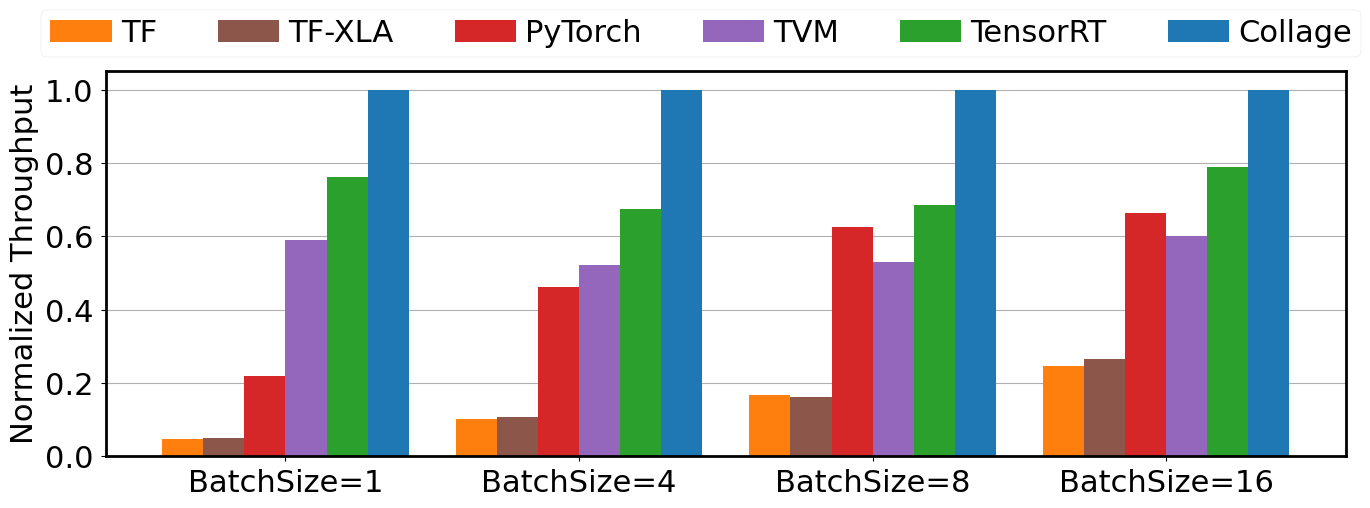}
	\caption{End-to-end performance with different batch sizes in ResNeXt50 on NVIDIA V100. Normalized throughput is the throughput normalized by the throughput of \ours{}. }
	\label{fig:e2e_perf_diff_batch_size}
\end{figure}

As the representative case, different batch sizes are also examined with ResNeXt50 on V100. Figure~\ref{fig:e2e_perf_diff_batch_size} indicates that \ours{} consistently outperforms the state-of-the-art frameworks across different batch sizes as well.

Since backends and their performance vary depending on the underlying execution environment, backend placement should be carefully customized by considering their performance landscape. Our experimental results indicate that \ours{} can stably offer a faster DL execution than existing frameworks with the rigid hand-written heuristics across different hardware architectures.


\subsection{Optimization Time}  \label{sec:opt_time}

To evaluate the overhead from our automated optimizer, this subsection studies the overall optimization cost of the two-level approach. For this section, we use NVIDIA V100 as our target.

\begin{figure}[t]
	\centering
	\includegraphics[width=0.9\linewidth]{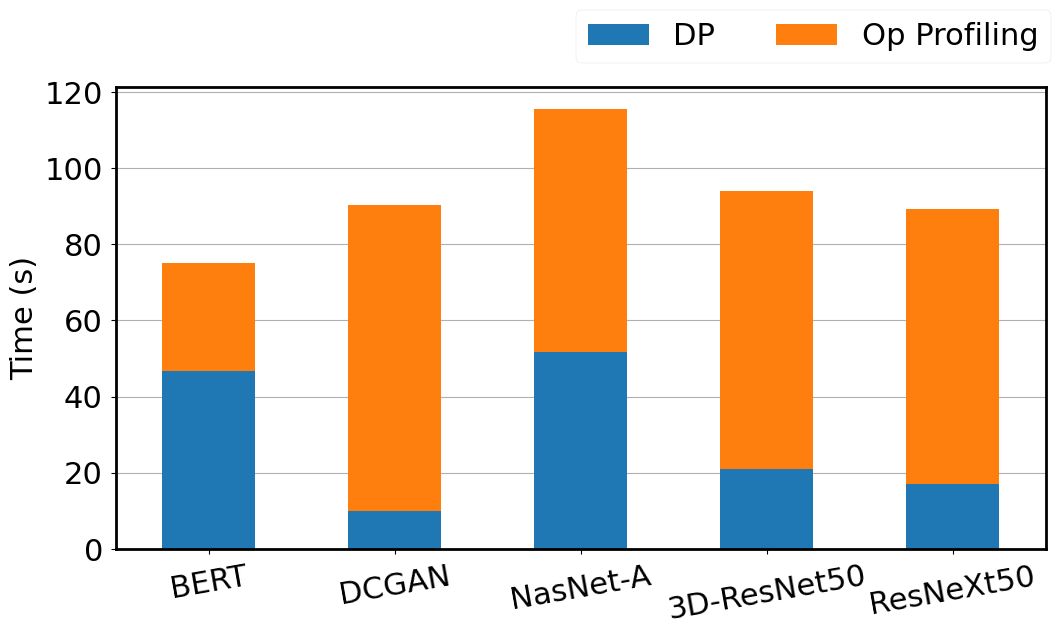}
	\caption{The breakdown of op-level placement optimization time. On average, profiling overhead for operator cost measurements takes up 68\% of the entire optimization time. Note that profiling is only necessary for unseen operators. Once the cost of a new operator is measured, its information will be saved in the logging database in \ours{} to avoid the repetitive profiling. If profiling log is available, op-level optimizer only takes less than a minute. }
	\label{fig:dp_tuning_time}
\end{figure}

Figure~\ref{fig:dp_tuning_time} shows the breakdown of our operator-level optimization time. If the optimization is launched from scratch, the entire optimization process takes up to two minutes. This optimization time consists of two parts: measurement of the operator cost and overhead from the DP algorithm. Due to the high evaluation cost, the optimization time is dominated by the profiling overhead. However, as discussed in \S\ref{sec:op-lv-tuner}, the repetitive profiling for operator cost can be avoided by saving the cost of each operator. When the cost of every operator is profiled in advance, our op-level placement optimization takes less than a minute on all of the five networks. 

\begin{figure}[t]
	\centering
	\includegraphics[width=0.9\linewidth]{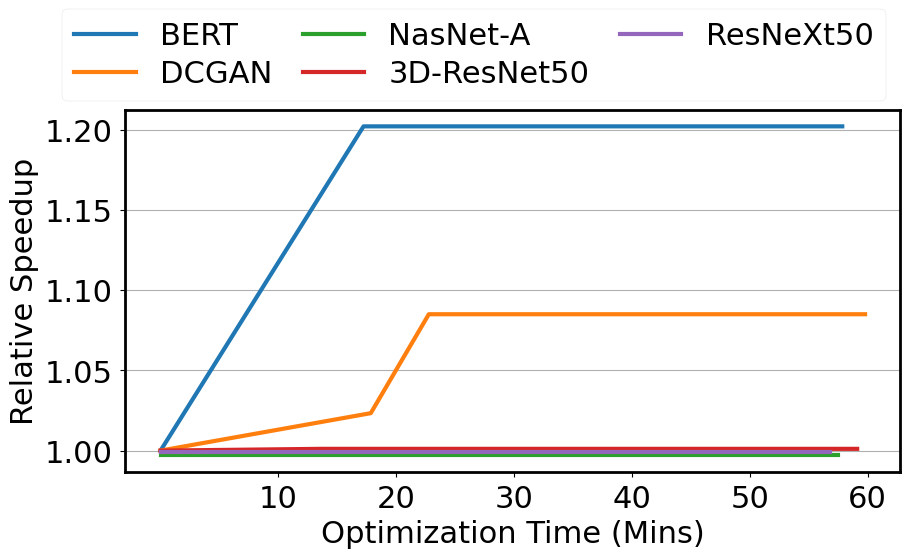}
	\caption{Performance improvement of graph-level placement optimization over time. The y-axis presents the speedup relative to the op-level placement optimization.}
	\label{fig:ev_tuning_progress}
\end{figure}

Figure~\ref{fig:ev_tuning_progress} exhibits how our graph-level placement optimization gradually improves from the op-level placement optimization over time. The evolutionary searcher could boost the performance by leveraging more cross-kernel optimizations as it goes through several generations of mutations and crossovers. In BERT and DCGAN, the effect of cross-kernel optimization is quite notable and thus, our graph-level placement optimizer accelerate its execution by $1.09-1.20 \times$ from the op-level optimization. For the rest of the workloads, graph-level placement optimization cannot improve any further since the placement from the op-level optimization is already hard to beat. Overall, most of workloads are observed to reach the saturation within thirty minutes. 

Due to the lack of the efficient cost model that can factor in the cross-kernel optimization effect, graph-level placement optimization has expensive evaluation overhead that leads to the longer optimization time compared to the op-level. Given that our op-level placement optimizer can identify high-performance backend placement for the most workloads within just a minute, we recommend the graph-level placement optimization as the optional tool for the users interested in squeezing the last drop of performance. 

\begin{figure*}[t]
	\centering
	\includegraphics[width=0.99\linewidth]{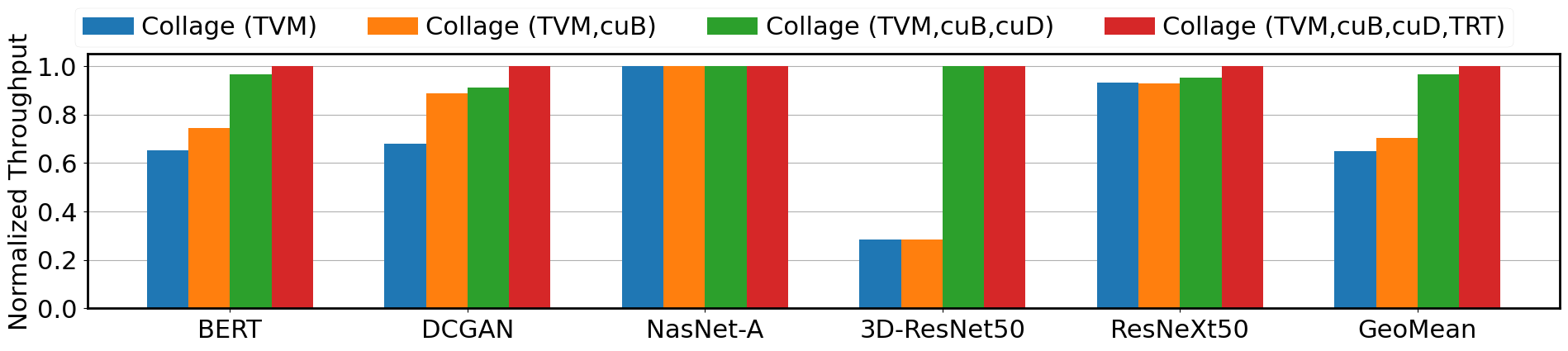}
	\caption{End-to-end performance of \ours{} with different number of backends on NVIDIA Tesla V100. Each throughput is normalized by the throughput of \ours{} (TVM,cuB,cuD,TRT). TVM, cuB, cuD, and TRT represents TVM, cuBLAS, cuDNN, and TensorRT.}
	\label{fig:backend_ablation}
\end{figure*}

\begin{figure}[t]
	\centering
	\includegraphics[width=0.99\linewidth]{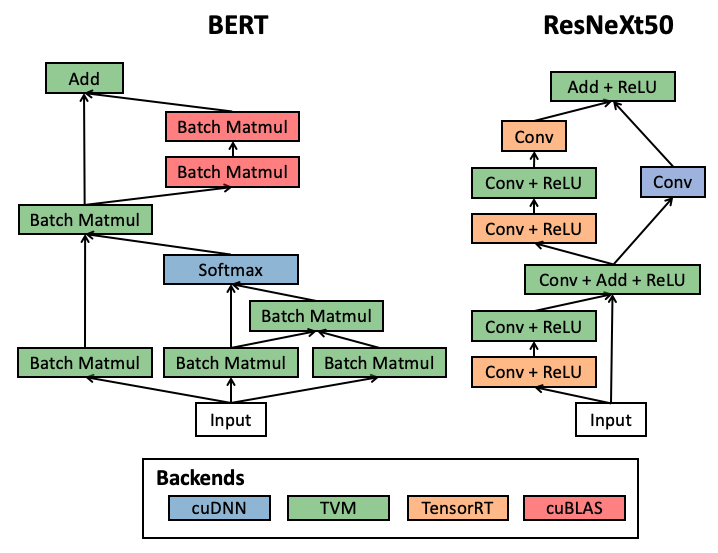}
	\caption{Representative backend placements discovered by \ours{} on V100 (\Cref{fig:e2e_perf_v100}). Note that \ours{} leverages various backends given their unique strength to enhance performance.}
	\label{fig:b_placement_example}
\end{figure}

\subsection{Backend Ablation Study} \label{sec:backend_ablation}

To assess the impact of integrating backends, we conduct an ablation study by adding backends one-by-one to \ours{}. 

\Cref{fig:backend_ablation} shows the experimental result on V100. Overall, \ours{} monotonically improves performance as we integrate more backends. This reinforces the importance of smart mixed-use of multiple backends and also corroborates the robustness of our backend placement optimization. It is worth noting that the performance improvement from a new backend varies depending on a network. In the case of BERT and DCGAN, we see relatively consistent enhancement from each backend. This is because \ours{} identifies a way to utilize every backend for the different part of the workload depending on its own unique strength.
In case of NasNet-A and ResNeXt50, TVM offers the majority of the performance improvement while cuDNN significantly benefits \ours{} for the 3D-ResNet50. 

These observations show that \ours{} can stably improve performance by having more backends. By leveraging the unique strength of available backends, our automated optimizer delivers the performance with a set of backends that surpasses or guarantees the performance with its subset.

\subsection{Case Study of Backend Operator Placement}  \label{sec:backend_case_study}
To understand the source of performance improvement from \ours{}, we examine two representative workloads in detail. Figure~\ref{fig:b_placement_example} illustrates \ours{}'s final backend placement for ResNeXt50 and BERT on V100. 

Even within a single network, we observe that the same type of operator is mapped to different backends due to the performance diversity depending on its configuration, such as data shape and kernel size, and the operator fusion with its neighbor nodes. For example, batch matrix multiplication operators in BERT are assigned to two different backends (cuBLAS and TVM) while convolution operators in ResNeXt50 are assigned to three different backends (cuDNN, TVM, and TensorRT). Interestingly, the graph inference library (e.g., TensorRT) can be a competitive choice even for a single operator as observed with some convolution operators in ResNeXt50.

This figure also demonstrates that \ours{} is capable of leveraging various fusion patterns from each backend. For instance, we discover a variety of operator fusion patterns selected by \ours{} such as Conv+ReLU, Conv+Add+ReLU, and Add+ReLU. Although it is omitted from this figure for simplicity, we observe the fusion pattern involved with more than ten operators. Again, as in a single operator, \ours{} chooses the different backends for the identical fusion pattern of Conv+Relu in ResNeXt50 because the best backend choice varies depending on specific operator configurations.

This study confirms that \ours{} can accelerate DL workload execution by leveraging diverse operator patterns from multiple backends given their performance characteristics.



	\section{Related Work}






\textbf{Diversified Backend Ecosystem.}
To extract the best performance from the underlying hardware, there have been substantial efforts to design high-performance DL backends. Hardware vendors have released various specialized optimized libraries and inference engines. NVIDIA has actively developed cuDNN~\cite{cuDNN} to deliver optimized implementations of DL operators, cuBLAS~\cite{cuBLAS} to offer efficient BLAS kernels, and TensorRT~\cite{TensorRT} to create fast execution plans for DL workloads. Particularly, TensorRT considers various graph-wide cross-kernel optimizations for scheduling, memory footprint and etc. Meanwhile, Intel has released oneDNN~\cite{OneDNN} for optimized DL operator kernels and OpenVINO~\cite{OpenVINO} as an inference engine for Intel CPUs. AMD also has driven MIOpen~\cite{khan2019miopen}, an open source GPU library for DL primitives.

Today's DL frameworks exploit tensor compilers~\cite{ragan2013halide, Halide_AutoSched, chen2018tvm, baghdadi2019tiramisu,fegade2021cortex, Ansor, lattner2021mlir, kjolstad2017tensor, phothilimthana2021flexible, ma2020autohoot, chelini2020automatic, rasch2019generating, bastoul2022optimizing, jeong2021union, truong2016latte, vasilache2018tensor, jung2021deepcuts, chandrasekhar2019igc, lueh2021c, ansel2014opentuner, zheng2020flextensor} as their backends to generate operator kernels for various target devices. While some tensor compilers rely on manual scheduling \cite{ragan2013halide, baghdadi2019tiramisu, zheng2020flextensor}, automatic approaches \cite{chen2018tvm, chen2018learning, jung2021deepcuts, Halide_AutoSched, Ansor, phothilimthana2021flexible, ragan2013halide, kjolstad2017tensor, chelini2020automatic, ma2020autohoot, jung2021deepcuts, rasch2019generating, bastoul2022optimizing, ansel2014opentuner, vasilache2018tensor} has been actively studied to optimize tensor operator kernels for a given DL workload and device. For instance, Tensor Comprehension~\cite{vasilache2018tensor} uses black-box auto-tuning to optimize CUDA kernels along with polyhedral optimizations. 
To speed up the optimization time, cost model has been also widely examined together with automated approaches~\cite{chen2018learning, kaufman2019learned, Ansor, zheng2020flextensor}.

By providing an expressive registration interface and automatic placement optimizer, \ours{} enables seamless integration of a wide variety of DL backends without any expertise in complex performance dynamics of varied backends.


\textbf{DL Frameworks.} To provide easy and powerful platform of running a variety of DL workloads, different frameworks have been continuously released and improved. Google maintains TensorFlow~\cite{TensorFlow} and XLA~\cite{XLA} to optimize the execution on various hardware devices including TPUs~\cite{TPU}. Facebook develops Pytorch~\cite{PyTorch} that supports dynamic eager execution for usability while preserving compelling DL execution performance. For NVIDIA GPUs, TensorRT~\cite{TensorRT} is developed as a runtime framework that optimizes DL model execution. As an open-source C++ library and compiler suite for CPUs, Intel has launched nGraph~\cite{cyphers2018intel}. Also, TVM~\cite{chen2018tvm} offers the efficient compilation pipeline that is designed to support diverse hardware devices and DL workloads. On the other hand, Glow~\cite{rotem2018glow} is proposed to efficiently generate the optimized code for multiple targets of heterogeneous hardware. While such existing DL frameworks employ handwritten rules to integrate new backend, \ours{} reduces the manual effort with the backend pattern abstraction and extracts further performance gain with the automated backend placement.



\textbf{Operator Fusion.} Fusion is one of the most efficient techniques to optimize DL workloads by combining multiple high-level operators on the computation graph into a single kernel. To maximize the benefit, advanced fusion techniques \citep{ma2020rammer, fegade2021cortex, niu2021dnnfusion, zheng2020fusionstitching, XLA, elgamal2017spoof, boehm2018optimizing, ashari2015optimizing, chen2018tvm, jung2021deepcuts, li2022automatic, diamos2016persistent, abdolrashidi2019learning, sivathanu2019astra} introduce their own unique fusion rules to apply this optimization beyond a few special cases. For instance, by iterating over every operator, TVM seeks for an opportunity to merge each operator with its neighbors by using the union-find algorithm~\cite{chen2018tvm}.
To efficiently explore the fusion opportunities, DNNFusion~\cite{niu2021dnnfusion} employs a detailed classification of operation type and makes the fusion decisions. To identify the best fusion plan, FusionStitching~\cite{zheng2020fusionstitching} conducts Just-In-Time tuning. NVIDIA has actively improved the fusion engine in cuDNN to merge certain patterns of operators at runtime~\cite{cuDNN}. Internally, TensorRT~\cite{TensorRT} also actively apply the fusion to optimize the memory access and scheduling overhead. By offering the highly flexible user interface for the pattern rules, \ours{} can support such complicated fusion patterns from a variety of such backends. With fusion patterns and their rules, \ours{} naturally considers diverse fusion possibilities in multiple backends.

\textbf{Graph Rewriting.} To accelerate a DL execution, DL frameworks can rewrite an input computation graph by considering a number of graph substitution rules. Most DL frameworks such as TensorFlow~\cite{TensorFlow}, TensorRT~\cite{TensorRT}, and TVM~\cite{chen2018tvm} rely on the greedy approach by opportunistically applying a few important hand-coded rules. In contrast, MetaFlow~\cite{jia2019optimizing} suggests an automated graph rewriting approach that optimizes an input graph using backtracking search. TASO~\cite{jia2019taso} extends MetaFlow's backtracking search and further automates graph substitution generation for every new input graph. To further improve graph substitution search efficiency, sampling-based approach \cite{fang2020optimizing} has also been explored. To overcome the inefficiency in making sequential rewriting decisions, \cite{yang2021equality} proposes e-graph and equality saturation method. As these graph rewriting techniques are orthogonal to \ours{}, \ours{} can improve the performance of a rewritten computation graph by optimizing the backend placement. 

\textbf{Device Placement.} There are two major categories of work that investigates how to place DL operators across devices. One category is to learn a placement policy~\cite{mirhoseini2017device, mirhoseini2018hierarchical, gao2018spotlight} that places each operator onto one of given set of devices and generalize it to new workloads via transfer learning \cite{zhou2019gdp, addanki2018placeto, Paliwal2020Reinforced}. Another category is to algorithmically find good graph partitions of DL workloads and their schedules \cite{FlexFlow, jia2018exploring, narayanan2019pipedream, tarnawski2020efficient, zheng2022alpa}; for example, FlexFlow~\cite{FlexFlow} uses stochastic search method with delta simulation to partition a single operator into multiple computation and place them on devices. Compared to device placement, backend placement itself has its unique challenges of modeling complicated and fast-evolving operator fusion patterns and constraints from diverse backends in addition to different backend characteristics (e.g., cross-kernel optimization of graph inference library). To tackle this challenge, \ours{} provides an expressive backend pattern abstraction and a two-level optimizer, each level of which considers different characteristics of backends. Our work is complementary to existing device placement works.

	\section{Conclusion}

This work investigates an efficient DL backend integration system, called {\em Collage}. For the seamless integration of various backends, Collage offers an user interface that allows the flexible specification of diverse backend capabilities. To find the best uses of available backends, Collage introduces a two-level optimization method and automatically customizes the best possible backend placement for the underlying execution environment. The experimental results demonstrate that \ours{} outperforms the best manual approach in the state-of-the-arts DL framework by up to $1.43\times$ on average over real-life DL models and various hardware architectures. More importantly, unlike existing approaches, it offers stable performance across diverse hardware architectures and models by selecting the most beneficial backends for each part of workload.


\section*{Acknowledgement}
We would like to thank members of Catalyst group at CMU for their helpful comments on our work and manuscript. We would also like to thank the anonymous PACT reviewers for constructive feedbacks. This work was partially supported by the National Science Foundation under grant number CNS-2147909 and the Real Time Machine Learning (RTML) DARPA project.

	\bibliographystyle{ACM-Reference-Format}
	\bibliography{reference}
	
\end{document}